\documentclass[10pt,twocolumn,letterpaper]{article}

\usepackage[pagenumbers]{cvpr} 

\usepackage{graphicx}
\usepackage{amsmath}
\usepackage{amssymb}
\usepackage{wasysym}
\usepackage{booktabs}

\usepackage[pagebackref,breaklinks,colorlinks]{hyperref}

\usepackage[capitalize]{cleveref}
\crefname{section}{Sec.}{Secs.}
\Crefname{section}{Section}{Sections}
\Crefname{table}{Table}{Tables}
\crefname{table}{Tab.}{Tabs.}

\def\cvprPaperID{17} 
\def\confName{CVPR}
\def\confYear{2022}

\newif\ifdraft
\drafttrue

\newcommand{\comment}[1]{}

\newcommand{\parag}[1]{{\vspace{2pt}\noindent\bf{#1}}}

\newcommand{\vk}{\mathbf{k}}

\newcommand{\vp}{\mathbf{p}}

\newcommand{\vw}{\mathbf{w}}

\newcommand{\vz}{\mathbf{z}}

\newcommand{\mH}{\mathbf{H}}

\begin{document}

\title{LatentKeypointGAN: Controlling Images via Latent Keypoints \\ Extended Abstract}

\author{Xingzhe He \hspace{5mm} Bastian Wandt \hspace{5mm} Helge Rhodin \\
University of British Columbia\\
{\tt\small \{xingzhe, wandt, rhodin\}@cs.ubc.ca}}

\maketitle

\begin{abstract}
Generative adversarial networks (GANs) can now generate photo-realistic images.
However, how to best control the image content remains an open challenge.
We introduce LatentKeypointGAN, a two-stage GAN internally conditioned on a set of keypoints and associated appearance embeddings providing control of the position and style of the generated objects and their respective parts. 
A major difficulty that we address is disentangling the image into spatial and appearance factors with little domain knowledge and supervision signals. 
We demonstrate in a user study and quantitative experiments that LatentKeypointGAN provides an interpretable latent space that can be used to re-arrange the generated images by re-positioning and exchanging keypoint embeddings, such as generating portraits by combining the eyes, and mouth from different images. 
Notably, our method does not require labels as it is self-supervised and thereby applies to diverse application domains, such as editing portraits, indoor rooms, and full-body human poses.
\end{abstract}

\begin{figure}
\centering
   \includegraphics[width=0.98\linewidth]{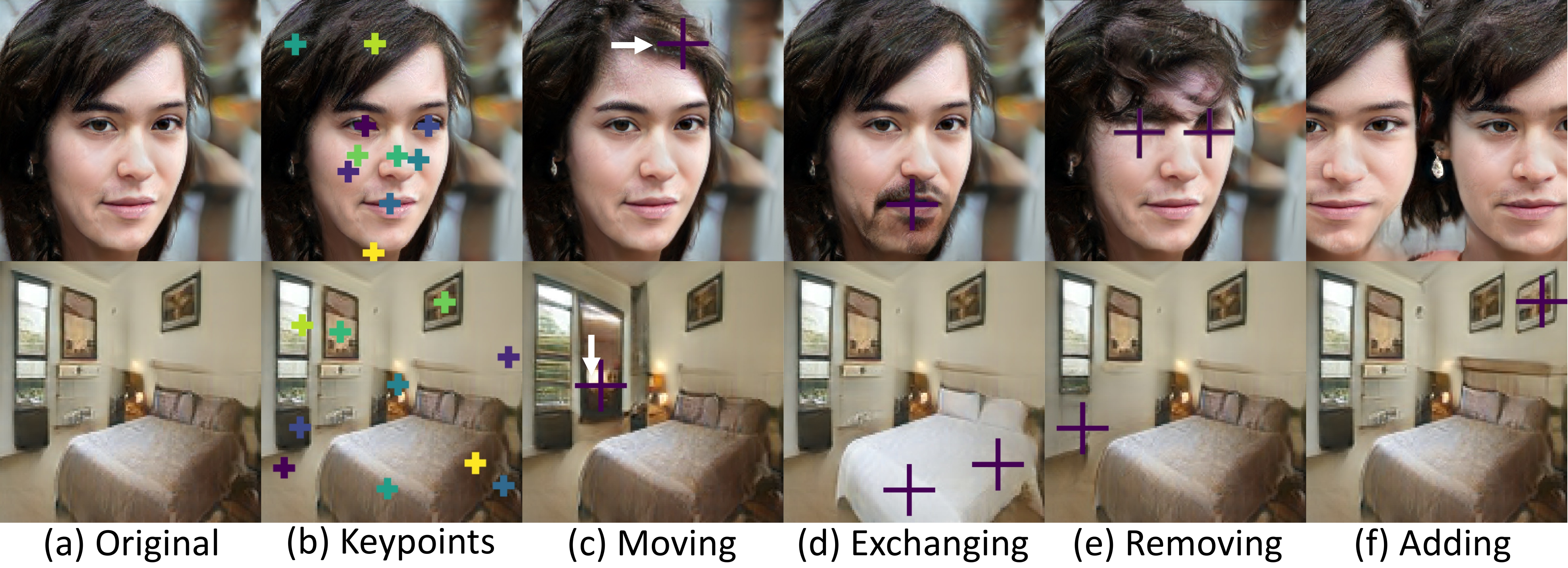}
\caption{GANs can generate phot-realistic images (a) but lack local editing capability. \textbf{LatentKeypointGAN} generates images with associated keypoints (a-b), which enables local editing by moving keypoints (c), exchanging appearance via the embedding we attach to each keypoint (d), removing parts by removing keypoints (e), and adding one or more parts by adding keypoints and their embedding (f). Our improvements are on the unsupervised learning of an interpretable latent space that disentangles pose and appearance, which makes it easy to use and applicable to diverse domains, including portraits (top row), indoor rooms (bottom row), and persons (see results section).}%
\label{fig:teaser}
\vspace{-10pt}
\end{figure}

\section{Introduction} \label{sec:intro}

While photo-realistic image generation is already reached for well-constrained domains, such as portraits, it remains challenging to make this synthesis interpretable and editable. Desired is a latent space that disentangles an image into parts and their appearances, which would allow a user to re-combine and re-imagine a generated portrait interactively and artistically. 

There are promising methods
\cite{thewlis2017unsupervised, zhang2018unsupervised, jakab2018unsupervised, lorenz2019unsupervised}
that use spatial image transformations
to create pairs of real and deformed images and impose an equivariance loss to discover keypoints and object appearance embeddings as the bottleneck of an autoencoder. 
Thereby, one can edit the image by moving the keypoints or modifying the appearance embedding. 
Yet, generated images lack fine detail. 
Methods such as StyleGAN~\cite{karras2019style, karras2020analyzing, abdal2019image2stylegan}, SPADE~\cite{park2019semantic} and their extensions \cite{collins2020editing, alharbi2020disentangled, kim2021exploiting, zhu2020sean} attain impressive quality and enable the mixing of appearance properties from different faces and to synthesize natural images by "painting" source and target regions in the feature space.
While powerful, their editing on feature maps makes it difficult to faithfully reposition image parts spatially. We show in Table~\ref{tab:features} the features that different methods have. Our goal is 
a user-friendly control via keypoints providing handles analogous to how character rigs are keyframed in classical animation.



Inspired by the control of autoencoder-based techniques and by the improved image quality of GANs, we introduce keypoint locations and associated feature embeddings as latent variables in the generator network of a GAN. Thereby, the location and appearance of image parts is separated and can be controlled. Figure~\ref{fig:teaser} shows how \emph{LatentKeypointGAN} enables editing the output image by changing the keypoint position, adding or removing points, and exchanging associated appearance embeddings locally while maintaining a high image quality that approaches that of existing GANs. 

We target an unsupervised setting in that the position, extend, and appearance of \emph{parts}---image regions that belong together and share appearance---is learned from unlabelled example images. It eases the application to new domains, where large image collections are available but exact segmentation masks or part labels are missing.

LatentKeypointGAN is designed as a two-stage GAN architecture that is trained end-to-end. In the first step, a generator network turns the input values sampled from a normal distribution into 2D keypoint locations and their associated encoding. We ensure with suitable neural network pathways that some of the encodings are correlated while others remain independent. These generated keypoints are then mapped to spatial heatmaps of increasing resolution. The heatmaps define the position of the keypoints and their support sets the influence range of their respective encodings. In the second step, a SPADE-like \cite{park2019semantic} image generator turns these spatial encodings into a complete and realistic image. 
Although entirely unsupervised, the learned keypoints meaningfully align with the image landmarks, such as a keypoint linked to the nose when generating images of faces, enabling the desired editing. 
\newcommand*\rot{\rotatebox{90}}

\begin{table}
\centering
\resizebox{\linewidth}{!}
{
\begin{tabular}[t]{ l r r r r r r r r r r}
Feature
& {\bf \rot{Zhang et al.~\cite{zhang2018unsupervised}}} 		
& {\bf \rot{Lorenz et al.~\cite{lorenz2019unsupervised}}} 		
& {\bf \rot{Karras et al.~\cite{karras2020analyzing}}} 		
& {\bf \rot{Collins et al.~\cite{collins2020editing}}} 	
& {\bf \rot{Alharbi et al.~\cite{alharbi2020disentangled}}} 		
& {\bf \rot{Kim et al.~\cite{kim2021exploiting}}} 		
& {\bf \rot{Wang et al.~\cite{wang2018high}}} 		
& {\bf \rot{Park et al.~\cite{park2019semantic}}} 		
& {\bf \rot{Zhu et al.~\cite{zhu2020sean}}}
& {\bf \rot{Ours} }		\\
\midrule
Appearance transfer	(global) & \CIRCLE & \CIRCLE	& \CIRCLE 	& \RIGHTcircle	& \CIRCLE		& \CIRCLE 		& \CIRCLE 		& \CIRCLE 		& \CIRCLE 	& \CIRCLE 		\\
Appearance transfer (local, part-based) & \CIRCLE & \CIRCLE  & \Circle 	& \RIGHTcircle 	& \CIRCLE 		& \CIRCLE 		& \CIRCLE 		& \CIRCLE 		& \CIRCLE 	& \CIRCLE 		\\
Removing and adding parts & \CIRCLE  & \CIRCLE  & \Circle 	& \Circle 	& \Circle 		& \Circle 		& \CIRCLE 		& \CIRCLE 		& \CIRCLE 	& \CIRCLE 		\\
Moving parts spatially 	 & \CIRCLE 	& \CIRCLE	& \Circle 	& \Circle 	& \Circle 		& \Circle 		& \Circle 		& \Circle 		& \Circle 	& \CIRCLE 		\\
Image quality w/o edits	& \Circle 	& \Circle & \CIRCLE &	\CIRCLE 	& \CIRCLE 		& \CIRCLE 		& \RIGHTcircle 		& \RIGHTcircle 		& \CIRCLE 	& \RIGHTcircle 		\\
Image quality after editing	& \Circle 	& \Circle  & \CIRCLE &	\CIRCLE 	& \CIRCLE 		& \CIRCLE 		& \RIGHTcircle 		& \RIGHTcircle		& \RIGHTcircle 	& \RIGHTcircle 		\\
Training w/o part annotation (unsupervised)	& \CIRCLE &	\CIRCLE   & \CIRCLE &	\CIRCLE 	& \CIRCLE 		& \CIRCLE 		& \Circle 		& \Circle 		& \Circle 	& \CIRCLE 		\\
Inference w/o manual feature region 'painting'	& \CIRCLE &	\CIRCLE & \CIRCLE &	\CIRCLE 	& \Circle 		& \Circle 		& \CIRCLE 		& \CIRCLE 		& \CIRCLE 	& \CIRCLE 		\\
\bottomrule
\end{tabular}
}
\caption{$\CIRCLE$ / $\RIGHTcircle$ / $\Circle$ : full / partial / no support; 
Feature table comparison to state-of-the-art generative image editing methods. %
For our task of intuitive editing, our method is a better fit than existing methods, though sacrificing some image quality over GANs with less control---an inevitable trade-off.
}
\label{tab:features}
\vspace{-10pt}
\end{table}
\section{Algorithm} \label{sec:method}

\begin{figure*}
\begin{center}
\includegraphics[width=0.95\linewidth]{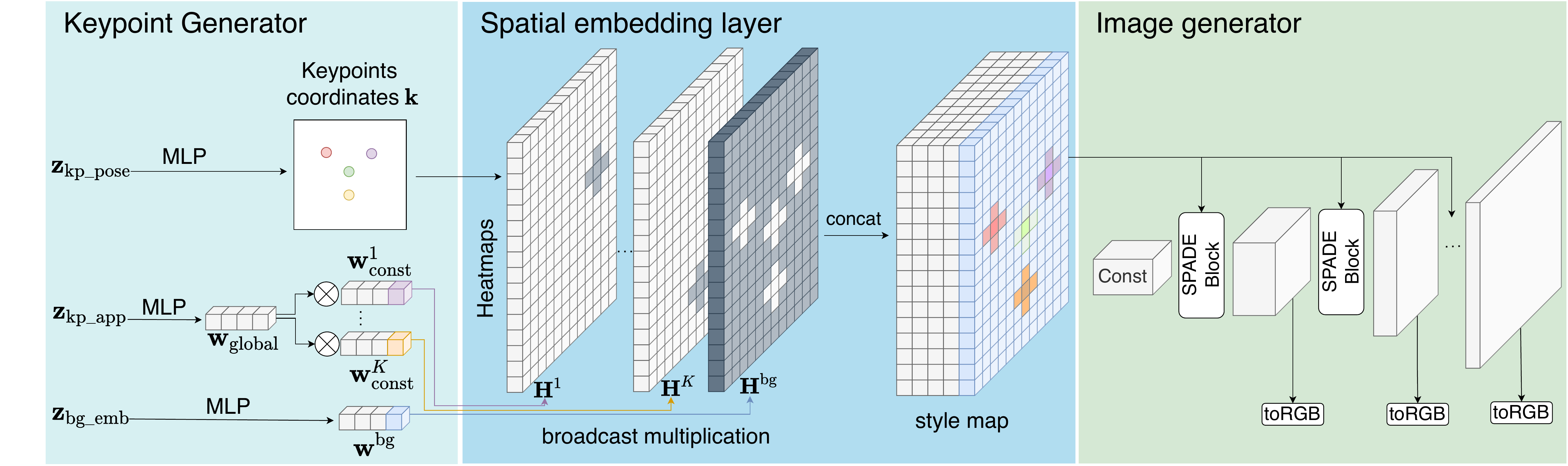}
\vspace{-10pt}
\end{center}
   \caption{\textbf{Overview.} Starting from noise $\vz$, LatentKeypointGAN generates keypoint coordinates, $\vk$ and their embeddings $\vw$. All keypoints are generated jointly. Crucial is how they are turned into feature maps that are localized around the keypoints, forming conditional maps for the image generation via SPADE block at different resolutions. At inference time, the position and embedding of keypoints can be edited by the user to control the position and appearance of parts.
   }
\label{fig:archi}
\vspace{-10pt}
\end{figure*}

Given a set of example images, our aim is a GAN that starting from random noise generates new images of the same type, e.g., indoor or portrait, and provides control over meaningful image parts by parametrizing the image in terms of keypoint locations $\{\vp\}_{i=1}^K$ and their embeddings $\{\vw\}_{i=1}^K$, where $K$ is the number of keypoints. We operate in the unsupervised setting, these parameters are latent variables that are inferred from the example images without requiring labels such as segmentation masks. Instead of direct supervision, we impose the following high-level objectives, either as loss functions or by channeling the information flow with tailored neural network layers. The detailed architecture can be found in Figure~\ref{fig:archi}.
%

\emph{GAN objective.} A ResNet discriminator \cite{he2016deep, karras2019style} is trained to distinguish real from fake images. The generator is trained to fool this discriminator. In our setting, the generated images must look realistic after editing or mixing the position and embedding of keypoints.

\emph{Pose-appearance disentanglement.} To enable editing of parts independently, we equip every keypoint with a 2D position $\vp^i$ and a feature embedding $\vw^i$, which are generated from independent noise $\vz_\text{kp\_pose}$ and $\vz_\text{kp\_app}$, respectively---either component can be edited independently. 

\emph{Part disentanglement.} The two eyes in a face naturally form separate parts, but we have to be able to model their differences (left-right orientation) and their symmetry (same eye color or glasses). We therefore split keypoint embeddings into constant independent components $\{\vw^i_\text{const}\}_{i=1}^K$ and a shared component $\vw_\text{global}$ that is generated from an independent noise vector $\vz_\text{kp\_app}$---providing independent user control at inference time.

\emph{Locality objective.} Keypoints should have a local support, influencing only nearby pixels. We use Gaussian heatmaps $\{\mH^i\}_{i=1}^K$ as intermediate for turning keypoints $\vp$ and their embedding $\vw$ into feature maps with a limited range of influence, with its smooth form enabling variations in scale---a local but learnable support.

\emph{Translation of parts.} Pixels around a keypoint---a part of the object---should move together under editing. This is imposed by using translation-invariant convolutional layers on the localized heatmaps, and pixel-wise operation SAPDE \cite{park2019semantic} on the localized feature maps.

\emph{Background separation.} Most images can be separated into foreground and background. To model this, we introduce a separate background "part" that controls those image regions not covered by keypoints and associated parts. This is generated from another independent noise $\vz_\text{bg\_emb}$.

\section{Results and Validation}   \label{experiments}
\paragraph{Benchmarks.} We use the official test splits from five different datasets. 
For \textbf{portrait editing} we use: a resolution of $512\times 512$ on FFHQ~\cite{karras2019style} to compare to GAN approaches; $256\times256$ on CelebA-HQ~\cite{zhu2020sean} to match FID comparison with the imge translation methods, including~SEAN; and $128\times 128$ on CelebA~\cite{liu2015faceattributes} for the autoencoder methods.
For both \textbf{human pose} experiments on BBCPose \cite{charles2013domain} and the \textbf{indoor} domain on LSUN Bedroom \cite{yu2015lsun} the resolution is $128\times 128$.Unless specified otherwise, we use 10 keypoints.

\parag{Image quality.} We use the FID \cite{heusel2017gans} for generation quality and the $\text{FID}_\text{lerp}$ \cite{kim2021exploiting} to measure global editing.




 
\begin{figure}[t!]
\begin{center}
\includegraphics[width=0.99\linewidth]{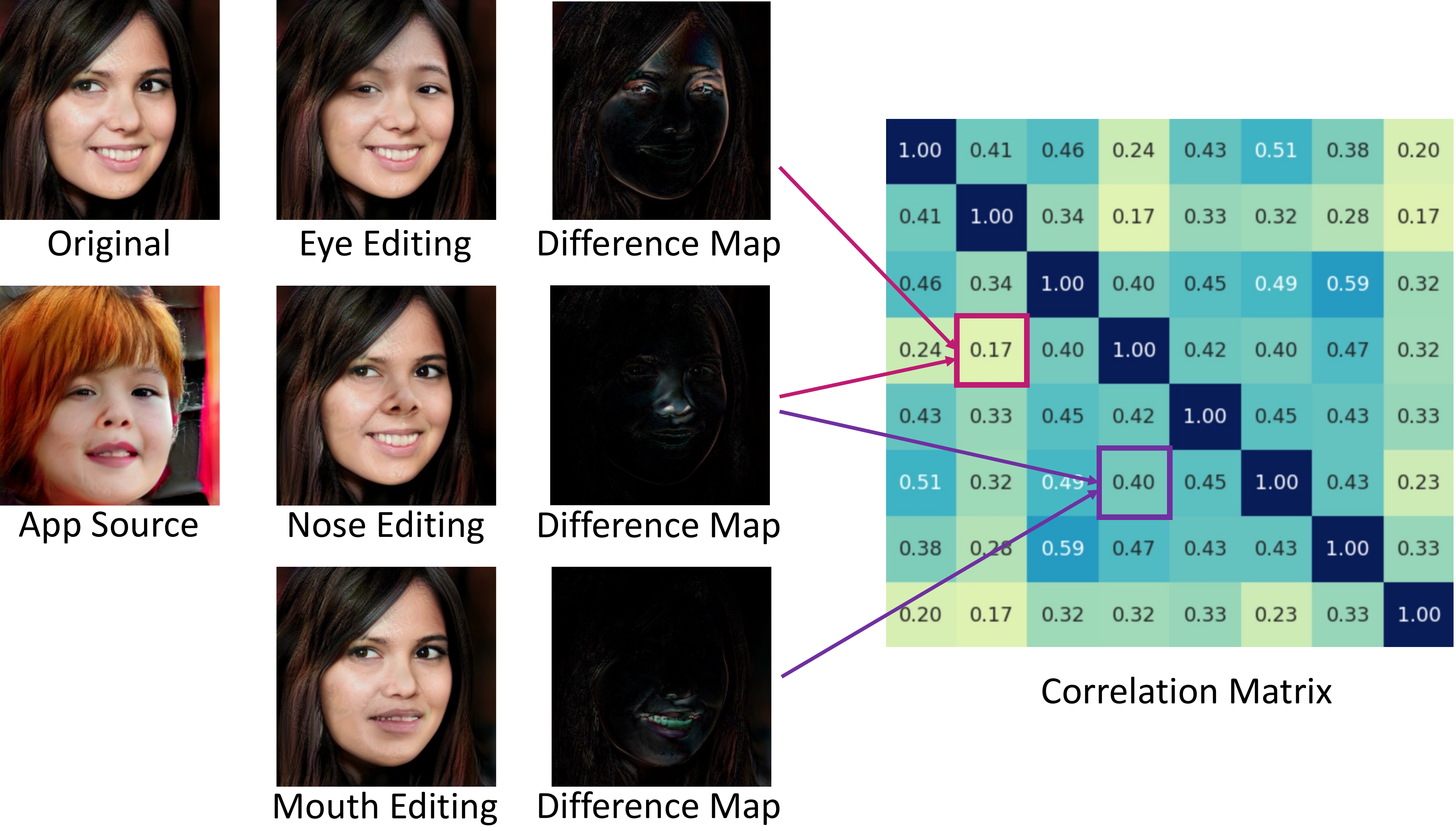}
\vspace{-10pt}
\end{center}
  \caption{\textbf{Our Correlation-based part Disentanglement (CPD) metric} is computed as the sum over part correlations. From a pair of images (left) the appearance of parts is exchanged one by one, here for eyes, nose, and mouth (center). The pairwise correlation of the resulting differences maps (right) forms the entries of the correlation matrix on which CPD is computed.}
\label{fig:CPD}
\vspace{-10pt}
\end{figure}

\parag{Disentanglement.} \label{sec:cpd} Previous disentangling methods analyzed latent space trajectories~\cite{karras2019style}, which does not generalize across network architectures, or use the difference magnitude between an original and edited image over manually annotated segmentation masks~\cite{collins2020editing}, which is not applicable to mask-free methods, including our keypoint approach. To this end, we propose a new correlation-based part disentanglement (CPD) that generalizes better. We generate 2000 images and randomly pair them. For a model with $K$ parts, we then create $K$ variants by exchanging part embeddings one by one. We take the difference before and after editing and compute the spatial correlation of parts over all 1000 pairs. 
Figure~\ref{fig:CPD} shows how the off-diagonal entries of the resulting correlation matrix quantify how much two different parts overlap.
Our CPD score captures this as one minus the average of the off-diagonal elements, which ranges from 1 (perfect disentanglement) to 0 (no control). For details, please refer to the supplemental document.

\subsection{Interactive Editing}

Our key advancement is the conditioning of a GAN on keypoint locations. Figure~\ref{fig:teaser} demonstrates how this enables moving of eyes and hair, and scaling of the entire face by scaling all keypoint positions. Surprisingly, due to the strong disentanglement of parts, the generator is not limited to the number of keypoints it is trained with. The teaser shows that removing and adding parts is possible by adding and removing an arbitrary number of peaks in the Gaussian heatmap and associated embedding masks. 
Figure~\ref{fig:functionality} compared editing capabilities to related GANs. None of these spatial operations have been demonstrated by existing GANs, as these focus on global and local editing via feature maps, but not on the repositioning and addition of parts as their parametrization on part position is not explicit. 



\newcommand{\centertable}[1]{\begin{tabular}{@{}c@{}} \vspace{-2cm}\\#1\\$\phantom{space}$ \end{tabular}}
\newcommand{\cna}{\centertable{n/a}}

\begin{figure}[t]
\huge
  \resizebox{0.98\linewidth}{!}{%
\begin{tabular}{cccccc@{}c@{}}
\centertable{\cite{kwon2021diagonal}}&
\includegraphics[width=0.35\linewidth]{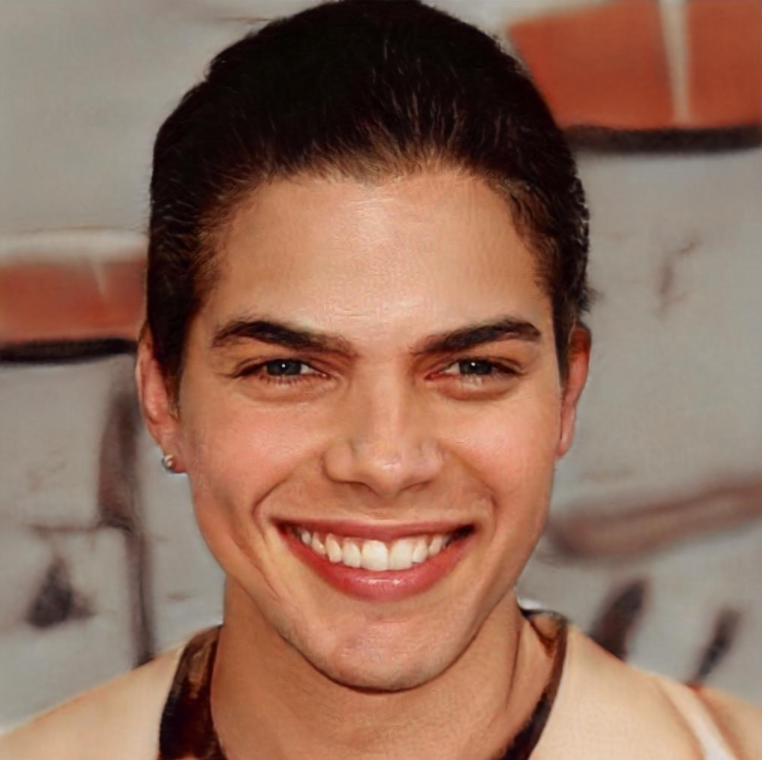}&
\includegraphics[width=0.35\linewidth]{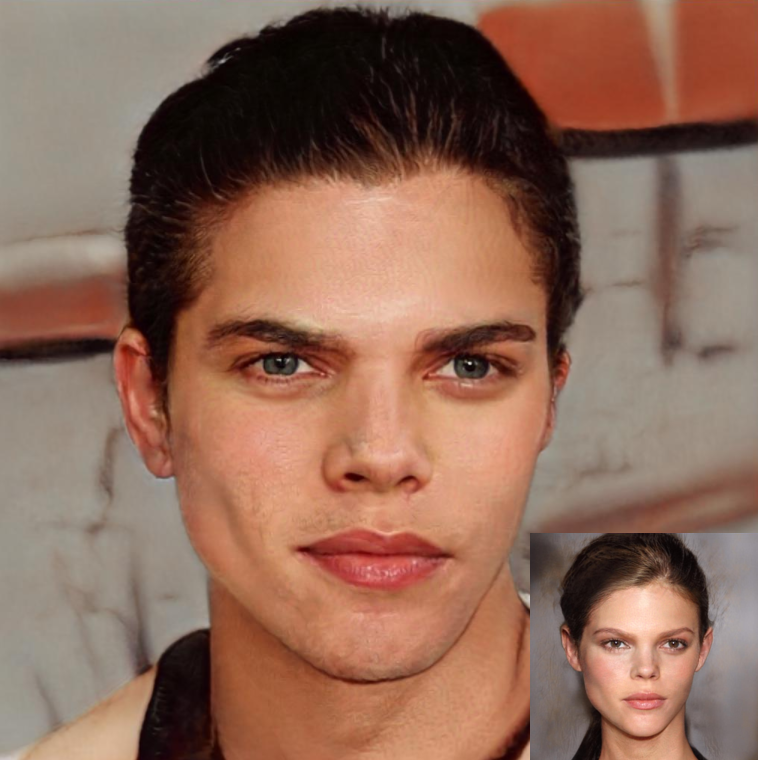}& \cna & \cna & \cna & \cna \\
\centertable{\cite{kim2021exploiting}}&
\includegraphics[width=0.35\linewidth]{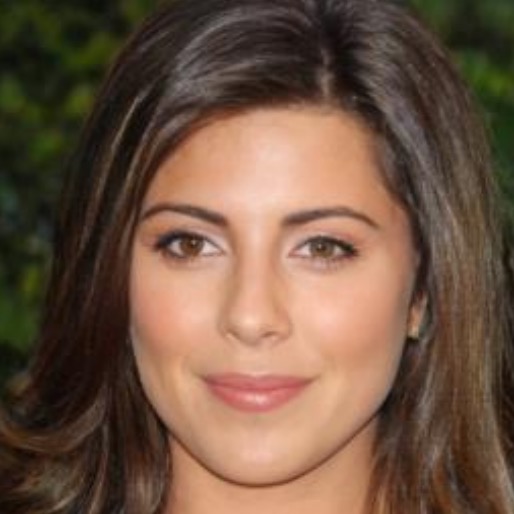}&
\includegraphics[width=0.35\linewidth]{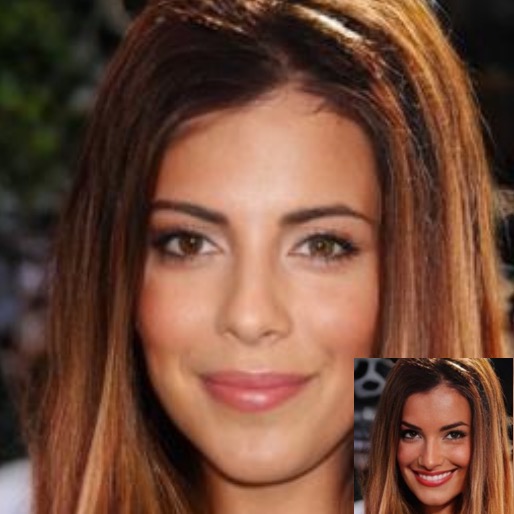}&
\includegraphics[width=0.35\linewidth]{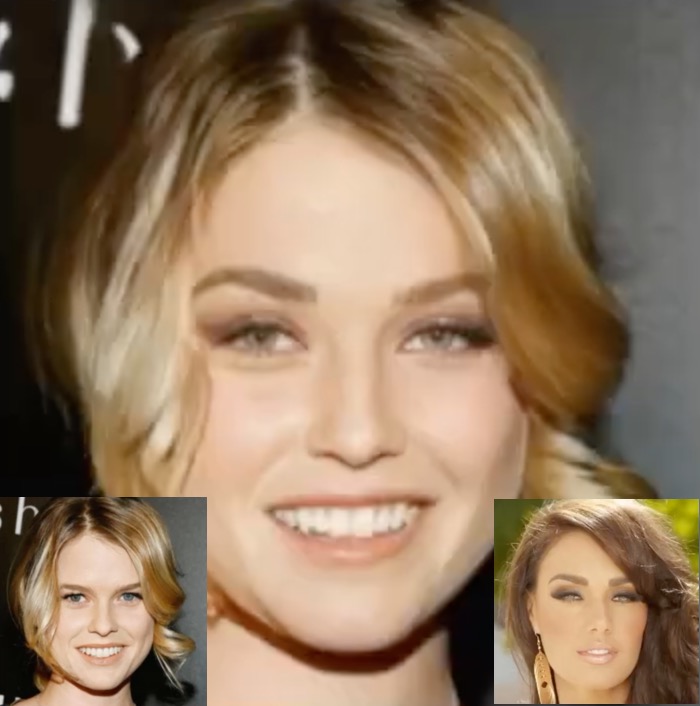}& \cna & \cna & \cna \\
\centertable{\cite{collins2020editing}}&
\includegraphics[width=0.35\linewidth]{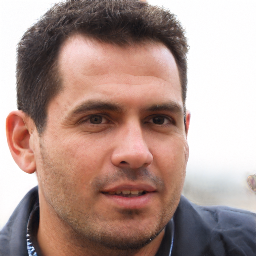}&
\includegraphics[width=0.35\linewidth]{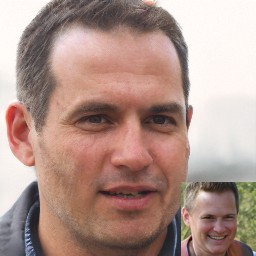}&
\includegraphics[width=0.35\linewidth]{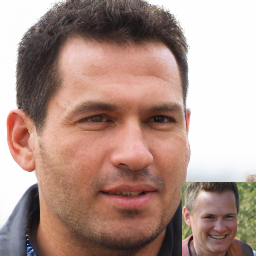}& \cna & \cna & \cna\\
\centertable{\cite{zhu2020sean}}&
\includegraphics[width=0.35\linewidth]{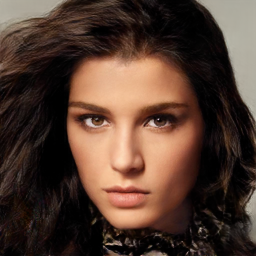}&
\includegraphics[width=0.35\linewidth]{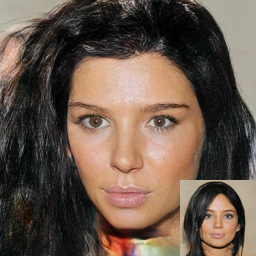}&
\includegraphics[width=0.35\linewidth]{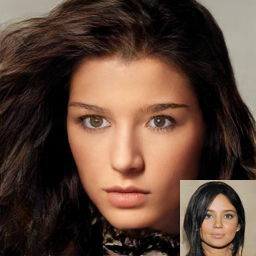}&
\includegraphics[width=0.35\linewidth]{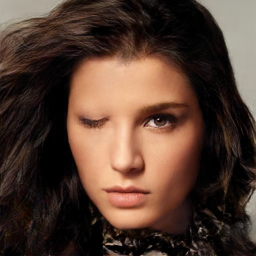}&
\includegraphics[width=0.35\linewidth]{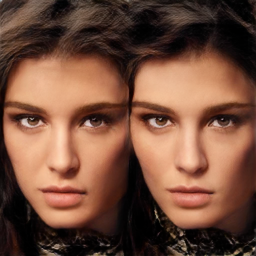}& \cna\\
\rot{$\phantom{ x:}$Ours}&
\includegraphics[width=0.35\linewidth]{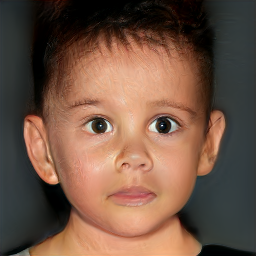}&
\includegraphics[width=0.35\linewidth]{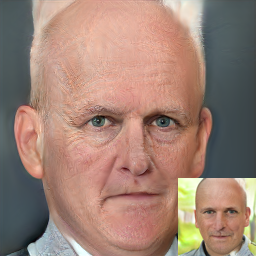}&
\includegraphics[width=0.35\linewidth]{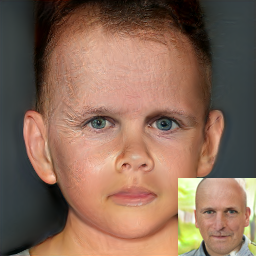}&
\includegraphics[width=0.35\linewidth]{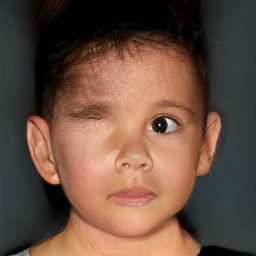}&
\includegraphics[width=0.35\linewidth]{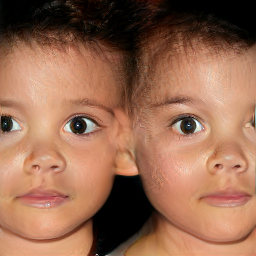}&
\includegraphics[width=0.35\linewidth]{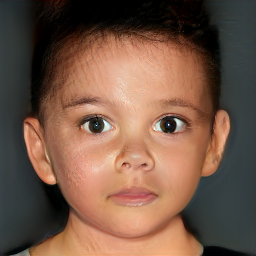}\\
 & Original & Global & Local & Removing & Adding & Moving\\
\end{tabular}}
\vspace{-5pt}
\caption{\textbf{Supported editing features.} While all of the tested methods enable global editing, only some offer local editing of part appearance, and non of the GAN-based methods (top three) demonstrated adding new parts, removing, nor animating parts via control handles, as we provide.
}
\label{fig:functionality}
\vspace{-10pt}
\end{figure}

\subsection{Editing Quality}
\label{sec:image_quality}

Our goal is to minimize the inevitable drop in image quality when enabling different levels of editing. 
For \textbf{global editing}, our $\text{FID}_\text{lerp}$ is close to that of all but the very recent approach by \cite{kim2021exploiting}, which however does not supports the desired positional editing.

\emph{User study.} Because there is no established metric for \textbf{local editing} and the autoencoder methods do not provide FID scores, we additionally conducted a comparative user study to  \cite{zhang2018unsupervised} (best autoencoder) on CelebA and \cite{zhu2020sean} (most disentangled prior work) on CelebA-HQ. Participants were asked to "Choose the image (A or B) with higher face quality" and "regardless of image quality" choose the pair that better preserves "facial features", "identity", and "outline of the face and its parts".
Compared to \cite{zhang2018unsupervised}, who supports the same keypoint editing as we do, our image quality is preferred drastically, by 92.17\%.
Also compared to, \cite{zhu2020sean} which reaches a better FID score before editing and better preserve the outlines with explicit segmentation masks (ours preferred 33.91\%, vs. 46.96\%), our face quality is preferred (by 94.78\%) after local editing mouth/eyes, and the identity preservation is nearly the same (49.57\% voted equal quality). We conclude that our edits are preferred because their masks and embeddings can become incompatible when exchanged while our keypoints are trained to be disentangled.
The full details are in the supplemental.

For \textbf{image generation} we are in the same ballpark as the supervised translation approaches (Table~\ref{tab:fid} FID, top half), yet above those unsupervised ones building upon StyleGAN (bottom half). This loosely quantifies the cost of imposing additional disentanglement constraints as very similar training strategies and generator architectures are used. 

\paragraph{Disentangled Representations.} \label{disentangled_representations}
The ability to move parts depends strongly on their disentanglement. 
We measure this as the overlap of the image regions controlled by pairs of parts using the CPD score explained in Figure~\ref{fig:CPD}. Table~\ref{tab:fid} compares CPD scores to all related methods that had trained models available. 
To compare to GANLocalEditing \cite{collins2020editing} with 8 parts, we grouped parts for SEAN and ours into semantically equivalent groups and, when comparing to GANLocalEditing, removed the background as not parametrized by them. 
Our method scores the best across all unsupervised methods (0.63 CPD vs. 0.45, 0.39, 0.35) and matches that of the supervised SEAN on their labeled dataset (0.7 CPD vs. 0.7), despite them having the advantage of conditioning on sharp segmentation masks while we only condition on self-supervised keypoints. This demonstrates the improvement in part disentanglement brought about by our contributions.

\begin{table}
\begin{center}
\resizebox{\linewidth}{!}{%
\begin{tabular}{|l|l|c|c|c|c|}
\hline
\multicolumn{2}{|l|}{Method} & Conditioned on & FID $\downarrow$ & $\text{FID}_\text{lerp}$ $\downarrow$ & CPD $\uparrow$\\
\hline
\multicolumn{2}{|l|}{Pix2PixHD \cite{wang2018high}} & masks & 23.69$\star$ & - & -\\ 
\multicolumn{2}{|l|}{SPADE \cite{park2019semantic}} & masks & 22.43$\star$ & - & -\\ 
\multicolumn{2}{|l|}{SEAN \cite{zhu2020sean}} & masks \& image & 17.66 & 30.29$\dagger$ & 0.70\\ \hline
\multicolumn{2}{|l|}{StyleGAN2 \cite{karras2020analyzing}} & unsupervised & 4.97 & 30.30$\dagger$ & -\\
\multicolumn{2}{|l|}{NoiseInjection \cite{alharbi2020disentangled}} & unsupervised & - & 27.96$\dagger$ & -\\
\multicolumn{2}{|l|}{StyleMapGAN \cite{kim2021exploiting}} & unsupervised & 4.72 & 9.97 & -\\
GANLocalEditing \cite{collins2020editing} & ($\epsilon=5$) & unsupervised & - & - & \textit{0.45}\\
 & ($\epsilon=50$) & unsupervised & - & - & \textit{0.39}\\
& ($\epsilon=500$) & unsupervised & - & - & \textit{0.35}\\
\multicolumn{2}{|l|}{Ours} & unsupervised & 16.26  & 31.80 & \textit{0.63}\\
\hline
\end{tabular}}
\end{center}
\caption{\textbf{Portrait image generation and editing quality} compared to supervised image translation on CelebA-HQ (top half) and to GANs on FFHQ (bottom). Our image quality scores before (FID column) and after editing  ($\text{FID}_\text{lerp}$) are close to the methods in each category, only outperformed by those providing fewer editing options or using supervision. This editing is enabled by the improved disentanglement (CPD column).
$\star$ is trained by \cite{zhu2020sean} and $\dagger$ by \cite{kim2021exploiting} using official implementations.
}
\label{tab:fid}
\end{table}

\paragraph{Generalization to Diverse Datasets.}

\begin{figure}[t]
\begin{center}
   \includegraphics[width=0.98\linewidth]{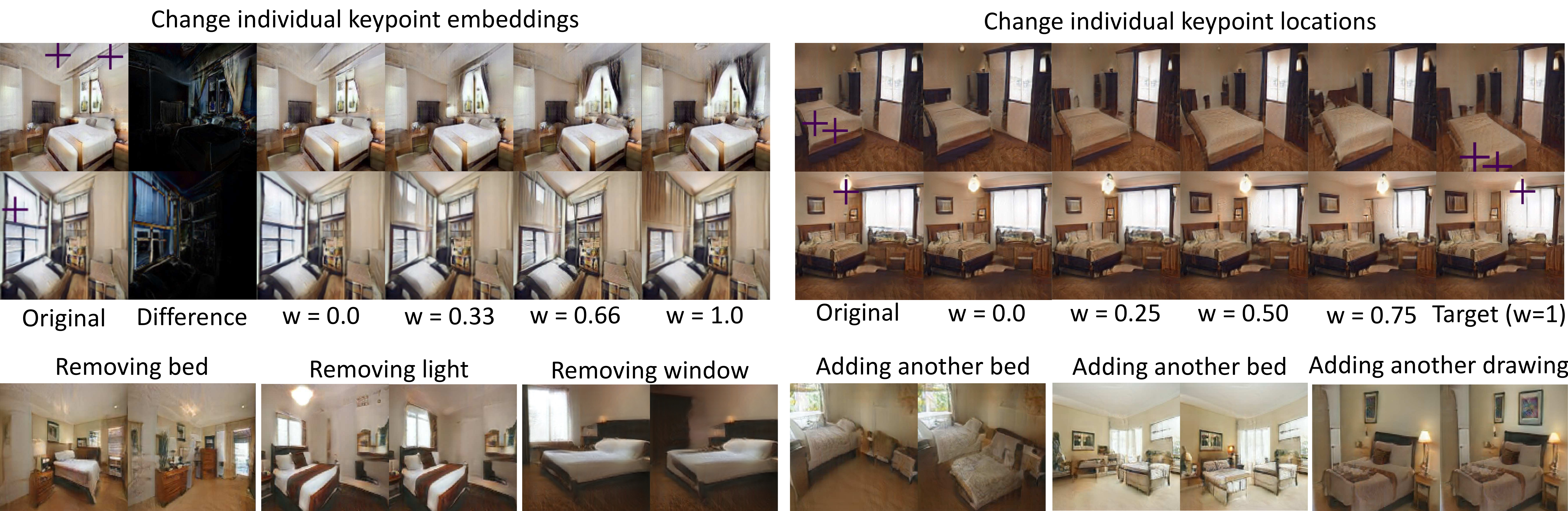}
   \vspace{-10pt}
\end{center}
   \caption{\textbf{Editing on Bedroom} by (top-left) interpolating the keypoint embeddings of curtain and window and (top-right) moving bed and light; and (bottom) removing and adding objects}
\label{fig:bedroom}
\end{figure}

\begin{figure}[t]
\begin{center}
   \includegraphics[width=0.98\linewidth]{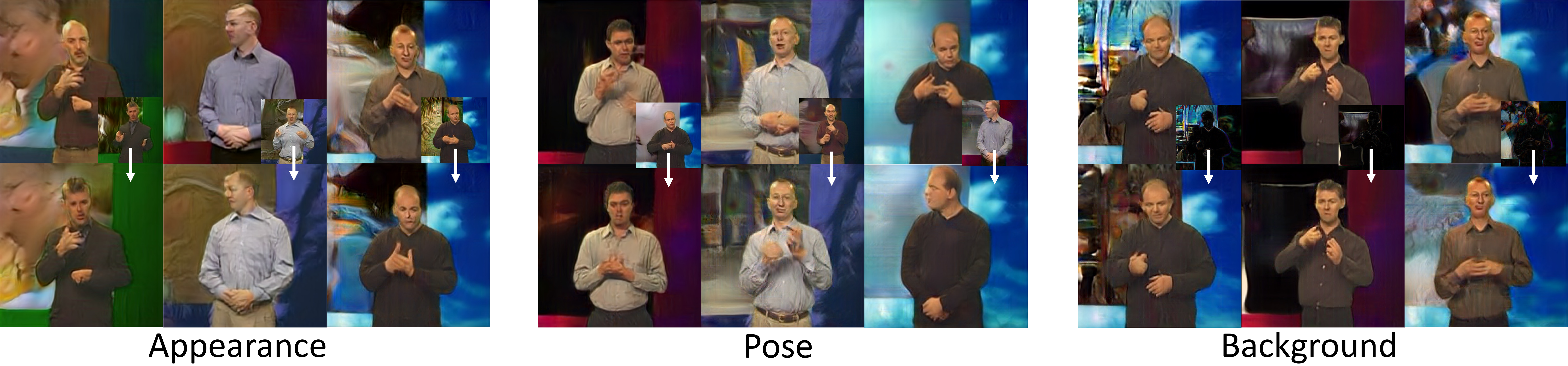}
   \vspace{-10pt}
\end{center}
   \caption{\textbf{Editing on BBC Pose.} The first row shows the source image and the second row the editing results. 
   \textbf{Left:} the human appearance is swapped with the small target image. 
   \textbf{Center:} changing the position to the one in the overlay. 
   \textbf{Right:} changing the background (inset shows the difference). 
   }
\label{fig:bbcpose}
\end{figure}

In Figure~\ref{fig:bedroom}, we explore the editing ability of entire scenes on the \textbf{LSUN bedroom dataset}. No previous unsupervised keypoint-based model has tried this difficult task before.
Figure~\ref{fig:bbcpose} explores the editing of \textbf{BBC Pose}. 
Although artifacts remain due to the detailed background and motion blur in the datasets, pose and appearance can still be exchanged.  

\section{Limitations and Future Work}
\label{sec:limitations}
While successful in disentangling facial details and objects in the bedroom dataset, the hair in portraits can mix with the background and, on the other hand, locally encoded features can lead to asymmetric faces, such as a pair of glasses with differently styled sides. If a different level of entanglement is desired, one could attempt to tune $\tau$ for each application domain or even each keypoint.
%
%
While the face orientation in portrait images can be controlled by moving keypoints, we found that orientation changes on the bedroom images are not reliable. We believe that it will be necessary to learn a 3D representation. 



{\small
\bibliographystyle{ieee_fullname}
\bibliography{egbib}

\begin{thebibliography}{10}\itemsep=-1pt

\bibitem{abdal2019image2stylegan}
Rameen Abdal, Yipeng Qin, and Peter Wonka.
\newblock Image2stylegan: How to embed images into the stylegan latent space?
\newblock In {\em Proceedings of the IEEE international conference on computer
  vision}, pages 4432--4441, 2019.

\bibitem{alharbi2020disentangled}
Yazeed Alharbi and Peter Wonka.
\newblock Disentangled image generation through structured noise injection.
\newblock In {\em Proceedings of the IEEE/CVF Conference on Computer Vision and
  Pattern Recognition}, pages 5134--5142, 2020.

\bibitem{charles2013domain}
James Charles, Tomas Pfister, Derek Magee, David Hogg, and Andrew Zisserman.
\newblock Domain adaptation for upper body pose tracking in signed tv
  broadcasts.
\newblock 2013.

\bibitem{collins2020editing}
Edo Collins, Raja Bala, Bob Price, and Sabine Susstrunk.
\newblock Editing in style: Uncovering the local semantics of gans.
\newblock In {\em Proceedings of the IEEE/CVF Conference on Computer Vision and
  Pattern Recognition}, pages 5771--5780, 2020.

\bibitem{he2016deep}
Kaiming He, Xiangyu Zhang, Shaoqing Ren, and Jian Sun.
\newblock Deep residual learning for image recognition.
\newblock In {\em Proceedings of the IEEE conference on computer vision and
  pattern recognition}, pages 770--778, 2016.

\bibitem{heusel2017gans}
Martin Heusel, Hubert Ramsauer, Thomas Unterthiner, Bernhard Nessler, and Sepp
  Hochreiter.
\newblock Gans trained by a two time-scale update rule converge to a local nash
  equilibrium.
\newblock In {\em Advances in neural information processing systems}, pages
  6626--6637, 2017.

\bibitem{jakab2018unsupervised}
Tomas Jakab, Ankush Gupta, Hakan Bilen, and Andrea Vedaldi.
\newblock Unsupervised learning of object landmarks through conditional image
  generation.
\newblock In {\em Advances in neural information processing systems}, pages
  4016--4027, 2018.

\bibitem{karras2019style}
Tero Karras, Samuli Laine, and Timo Aila.
\newblock A style-based generator architecture for generative adversarial
  networks.
\newblock In {\em Proceedings of the IEEE conference on computer vision and
  pattern recognition}, pages 4401--4410, 2019.

\bibitem{karras2020analyzing}
Tero Karras, Samuli Laine, Miika Aittala, Janne Hellsten, Jaakko Lehtinen, and
  Timo Aila.
\newblock Analyzing and improving the image quality of stylegan.
\newblock In {\em Proceedings of the IEEE/CVF Conference on Computer Vision and
  Pattern Recognition}, pages 8110--8119, 2020.

\bibitem{kim2021exploiting}
Hyunsu Kim, Yunjey Choi, Junho Kim, Sungjoo Yoo, and Youngjung Uh.
\newblock Exploiting spatial dimensions of latent in gan for real-time image
  editing.
\newblock In {\em Proceedings of the IEEE/CVF Conference on Computer Vision and
  Pattern Recognition}, pages 852--861, 2021.

\bibitem{kwon2021diagonal}
Gihyun Kwon and Jong~Chul Ye.
\newblock Diagonal attention and style-based gan for content-style
  disentanglement in image generation and translation.
\newblock In {\em Proceedings of the IEEE international conference on computer
  vision}, pages 13980--13989, 2021.

\bibitem{liu2015faceattributes}
Ziwei Liu, Ping Luo, Xiaogang Wang, and Xiaoou Tang.
\newblock Deep learning face attributes in the wild.
\newblock In {\em Proceedings of International Conference on Computer Vision
  (ICCV)}, December 2015.

\bibitem{lorenz2019unsupervised}
Dominik Lorenz, Leonard Bereska, Timo Milbich, and Bjorn Ommer.
\newblock Unsupervised part-based disentangling of object shape and appearance.
\newblock In {\em Proceedings of the IEEE Conference on Computer Vision and
  Pattern Recognition}, pages 10955--10964, 2019.

\bibitem{park2019semantic}
Taesung Park, Ming-Yu Liu, Ting-Chun Wang, and Jun-Yan Zhu.
\newblock Semantic image synthesis with spatially-adaptive normalization.
\newblock In {\em Proceedings of the IEEE Conference on Computer Vision and
  Pattern Recognition}, pages 2337--2346, 2019.

\bibitem{thewlis2017unsupervised}
James Thewlis, Hakan Bilen, and Andrea Vedaldi.
\newblock Unsupervised learning of object landmarks by factorized spatial
  embeddings.
\newblock In {\em Proceedings of the IEEE international conference on computer
  vision}, pages 5916--5925, 2017.

\bibitem{wang2018high}
Ting-Chun Wang, Ming-Yu Liu, Jun-Yan Zhu, Andrew Tao, Jan Kautz, and Bryan
  Catanzaro.
\newblock High-resolution image synthesis and semantic manipulation with
  conditional gans.
\newblock In {\em Proceedings of the IEEE conference on computer vision and
  pattern recognition}, pages 8798--8807, 2018.

\bibitem{yu2015lsun}
Fisher Yu, Ari Seff, Yinda Zhang, Shuran Song, Thomas Funkhouser, and Jianxiong
  Xiao.
\newblock Lsun: Construction of a large-scale image dataset using deep learning
  with humans in the loop.
\newblock {\em arXiv preprint arXiv:1506.03365}, 2015.

\bibitem{zhang2018unsupervised}
Yuting Zhang, Yijie Guo, Yixin Jin, Yijun Luo, Zhiyuan He, and Honglak Lee.
\newblock Unsupervised discovery of object landmarks as structural
  representations.
\newblock In {\em Proceedings of the IEEE Conference on Computer Vision and
  Pattern Recognition}, pages 2694--2703, 2018.

\bibitem{zhu2020sean}
Peihao Zhu, Rameen Abdal, Yipeng Qin, and Peter Wonka.
\newblock Sean: Image synthesis with semantic region-adaptive normalization.
\newblock In {\em Proceedings of the IEEE/CVF Conference on Computer Vision and
  Pattern Recognition}, pages 5104--5113, 2020.

\end{thebibliography}
}

\end{document}